\documentclass[10pt,twocolumn,letterpaper]{article}

\usepackage{times}
\usepackage{epsfig}
\usepackage{graphicx}
\usepackage{amsmath}
\usepackage{amssymb}
\usepackage{multirow}
\usepackage{booktabs}
\usepackage{array}

\usepackage[pagebackref=true,breaklinks=true,letterpaper=true,colorlinks,bookmarks=false]{hyperref}

\begin{document}

\title{Hi Sheldon! Creating Deep Personalized Characters \\ from TV Shows}

\author{Meidai Xuanyuan$^1$, Yuwang Wang$^2$, Honglei Guo$^2$, Xiao Ma$^1$,\\
Yuchen Guo$^3$, Tao Yu$^4$, Qionghai Dai$^2$ \\
Tsinghua University\\
{\tt\small $^1$\{xymd22, x-ma21\}@mails.tsinghua.edu.cn,}\\
{\tt\small$^2$\{wang-yuwang, guohonglei,  qhdai\}@mail.tsinghua.edu.cn,} \\
{\tt\small $^3$yuchen.w.guo@gmail.com, $^4$ytrock@tsinghua.edu.cn }
}
\date{}
\maketitle

\begin{abstract}
Imagine an interesting multimodal interactive scenario that you can see, hear, and chat with an AI-generated digital character, who is capable of behaving like {Sheldon} from {The Big Bang Theory}, as a DEEP copy from appearance to personality.  
Towards this fantastic multimodal chatting scenario, we propose a novel task, named Deep Personalized Character Creation (DPCC): creating multimodal chat personalized characters from multimodal data such as TV shows. 
Specifically, given a single- or multi-modality input (text, audio, video), the goal of DPCC is to generate a multi-modality (text, audio, video) response, which should be well-matched the personality of a specific character such as {Sheldon}, and of high quality as well. 
To support this novel task, we further collect a character centric multimodal dialogue dataset, named Deep Personalized Character Dataset (DPCD), from TV shows. 
DPCD contains character-specific multimodal dialogue data of ~10k utterances and ~6 hours of audio/video per character, which is around 10 times larger compared to existing related datasets. 
On DPCD, we present a baseline method for the DPCC task and create 5 Deep personalized digital Characters (DeepCharacters) from Big Bang TV Shows. 
%
We conduct both subjective and objective experiments to evaluate the multimodal response from DeepCharacters in terms of characterization and quality. The results demonstrates that, on our collected DPCD dataset, the proposed baseline can create personalized digital characters for generating multimodal response.
Our collected DPCD dataset, the code of data collection and our baseline will be published soon.  \footnote[1]{\href {https://github.com/Metaverse-AI-Lab-THU/Deep-Personalized-Character-Dataset-DPCD}{DPCD github repo.}}.

\end{abstract}

\section{Introduction}

\begin{figure}[pt]
  \centering
  \includegraphics[width=0.98\linewidth]{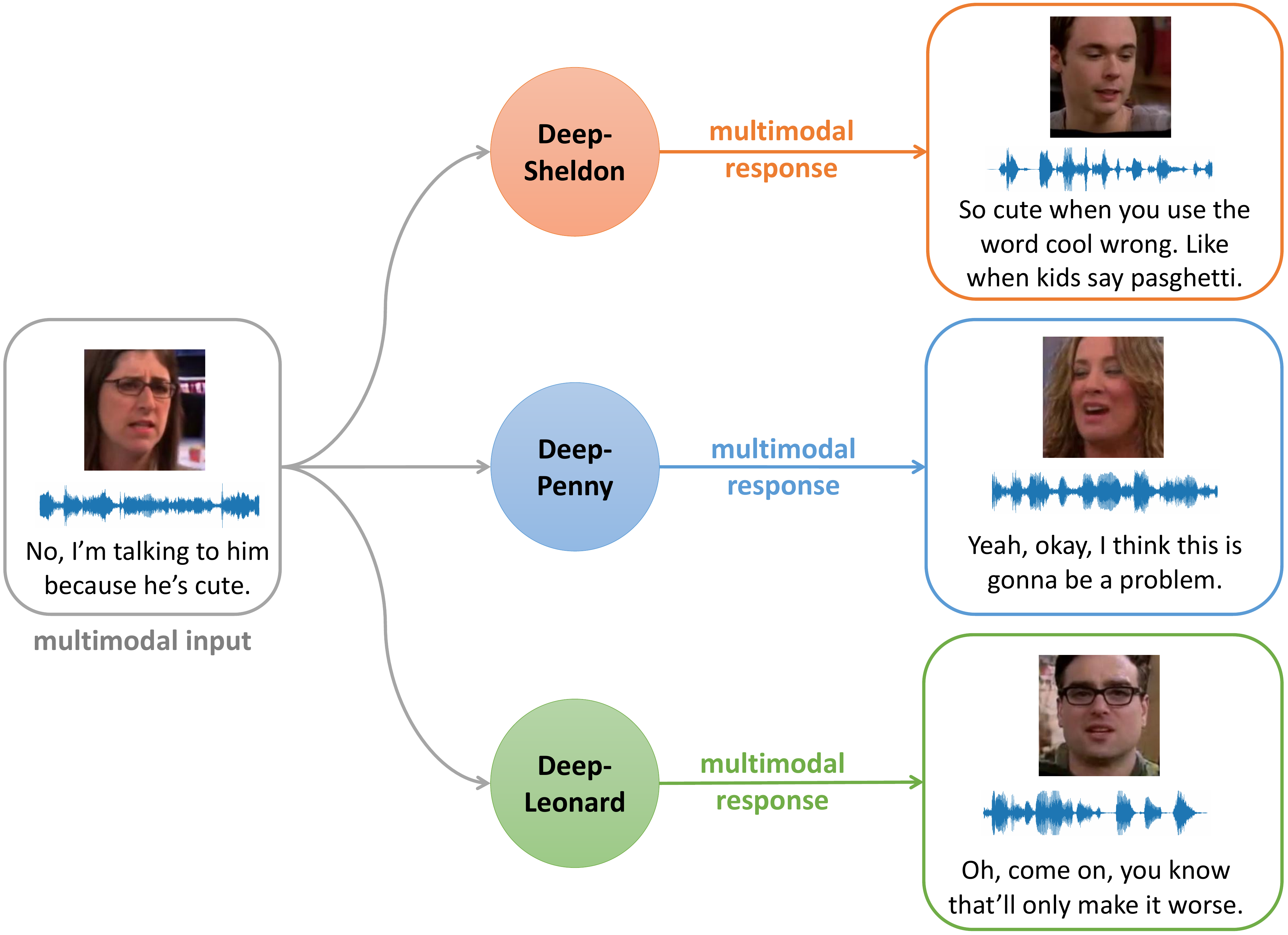}
   \caption{Our proposed Deep Personalized Character Creation (DPCC) task aims to create DeepCharacters, providing charaterized multimodal response, given the same stimuli.}
   \label{fig:teaser}
\end{figure}

Advancements in artificial intelligence have recently enabled highly realistic content synthesis, such as text to speech~\cite{ren2019fastspeech,arik2017deep,popov2021grad}, text to image~\cite{ramesh2021zero,rombach2021highresolution,saharia2022photorealistic,ramesh2022hierarchical}, text to 3D~\cite{poole2022dreamfusion}, text to video~\cite{singer2022make}, speech-driven facial animation~\cite{wu2021imitating,li2017learning,vougioukas2020realistic,siarohin2019first}, etc. 
These AI techniques lay the foundation of AI-generated characters, which have wide application in social services such as education, healthcare, and so on~\cite{pataranutaporn2021ai}.
We expect to interact with an AI-generated personalized character, just feels like interacting with a specific person, and the AI-generated character can respond to you with personalized text, voice and facial expression. 
Instead of creating a totally factitious character from manual settings, in this paper, we are heading for a novel setting: creating a \emph{DEEP} personalized specific character from his/her dialog videos providing textual, vocal and visual data.
By \emph{DEEP}, we mean not only mimicing the \emph{outer} characteristics such as appearance and tune, but also the \emph{inner} characteristics such as personality. 
Once we get a \emph{DEEP} personalized Sheldon (\emph{DeepSheldon}), one can have a multimodal chat with him and feel like we are really talking to the one you know from The Big Bang Theory. 
This technique of creating this kind of specific personalized digital characters has commercial demand in entertainment industry, such as generating virtual content of characters in TV shows, movies, games and concerts. 
One can imagine an interesting and exciting scenario that the AI-generated characters interact with each other and continue their story in a visual world line.

Towards this fantastic goal, we propose a novel task, named Deep Personalized Character Creation (DPCC), as generating a personalized character from the multimodal conversation data including text, audio, and video. 
As shown in Figure~\ref{fig:teaser}, DPCC takes the multimodal stimuli as input, and responds with a personalized multimodal output. 
Given the same input, different Deep personalized digital Characters (DeepCharacters) should provide responses that well reflect their individual trait, respectively. 
With multimodal response, DeepCharacters can bring more concrete and immersive interactive experience, compared to previous chatbots, which can only provide text or voice response. 
DPCC can also support text-only input and generate multimodal response.
Please note that this task is different from text or voice-driven generation, which is mainly converting from one modality to another~\cite{wu2021imitating,li2017learning,vougioukas2020realistic,siarohin2019first}. 
In our DPCC, the input is the stimuli, no explicit content of the response is provided, and each modality content of the response needs to be predicted. 
Also, different from previous personalized conversation generation tasks, our DPCC doesn't require additional profiles or traits as input to explicitly label personality, and learn the character's interaction pattern, speaking style and personality from conversation contexts.


To support DPCC task, we collect a multimodal conversation dataset, named Deep Personalized Character Dataset (DPCD). 
DPCD containing 5 main characters from \emph{The Big Bang Theory}\footnote{https://bigbangtrans.wordpress.com/}, and in total 29.75 hours videos and 29k conversation turns. 
We only collect the video clips containing characters' talking face to 
extract effective facial expression features. Besides, we manually check the auto-alignment between text and audio/video modality. We also conduct preprocessing to remove noise in audio. Those efforts ensure the collected data is of high quality for DPCC task.
There do exist multimodal (text, audio, video) datasets~\cite{ex5,ex6,ex7,ex8,ex9} for natural language processing (NLP), audio-visual synchronization (AVS) and speech synthesis. 
These existing multimodal datasets are collected for a general understanding of conversation content, emotion, or general synthesis of speech and talking head that can be applied to different persons, not for modelling specific characters. 
Consequently, the data for modeling the personality of a specific character is typically insufficient. For some datasets, the character labels of conversations are even not provided. 
As shown in Table~\ref{tab:With all other datasets} in Section \ref{subsec: dataset}, the average number of utterances for each character is at most 1,000 (on IEMOCAP ~\cite{ex5}), while our DPCD focus on the main characters and can provide almost an order more (9,718 utterances per character). 
Besides, those datasets are often of low quality in terms of text/audio/video misalignment, noisy audio, and face missing. 

 We present a baseline solution for DPCC task and create 5 DeepCharacters from the collected DPCD. 
To evaluate the textual and multimodal response from DeepCharacters in terms of characterization and quality, we design some tools to evaluate the characterization quality.  
Both subjective and objective experimental results demonstrate that the response of our created {DeepCharacters} can well reflect the personality of characters, indicating DPCD can be used for DPCC task.  

Our main contributions can be summarized as: 
$(i)$ We introduce a novel tasks to create deep personalized specific characters, who can generate the personalized multimodal response.  We believe this task can facilitate the research on multimodal and personalized AI generation.
$(ii)$ We collected a new dataset for the proposed task, which fills the gap of lacking high-quality, large-scale and character-centric conversation data. We think this dataset can benefit the community and step further towards personalized character creation. 
$(iii)$ We provide a baseline solution for this proposed task, and design some tools to evaluate the created DeepCharacters in terms of both characterization and quality. This lays the foundation of how to create and evaluate DeepCharacters for future works.

\section{Related Work}

Modeling personalized conversation has long been a complicated issue. Li et al.\cite{ex16} address the challenge of persona consistency by encoding personas and adding persona embeddings into seq2seq model. They generate persona-specific responses based on the conversations of 13 characters from TV series \emph{Friends} and \emph{The Big Bang Theory}. Based on the persona-based model by Li et al.\cite{ex16}, Xing et al.\cite{ex17} estimate the OCEAN scores for each speaker to train persona embedding.

Considering the difficulty and complexity of persona-based conversation modeling, many models rely on extra profile or personality traits to develop a sense of persona. Mazaré et al.\cite{ex18} define persona as a set of sentences representing the personality of the responding agent and train persona-based dialogue models on a large-scale persona-based dialogue dataset (PCR) extracted from REDDIT\footnote{https://www.reddit.com/r/datasets/
comments/3bxlg7/}. Zhong et al.\cite{ex19} further extend PCR with annotated empathy information. Zheng et al.\cite{ex20} build a persona-based empathetic conversation response selection model from amounts of multi-turn empathetic conversation data with five personality annotations (i.e.Gender, Age, Location, Interest Tags, and Self-description) for each speaker.  Wu et al.\cite{ex21} propose a generative split memory network to incorporate diverse personal information and generate personalized responses based on a personalized dataset PER-CHAT with finer-grained user personal information and contextual comments. Chen et al. \cite{chen2022cped} build the personality detection and emotion recognition model from a Chinese personalized dataset CPED with annotations of speakers' personalities, dynamic emotions and dialog actions.

Constrained by the lack of single speaker's conversation data and the complexity of personality modeling, the models listed above mainly focus on personal information consistency rather than character consistency, and rely heavily on detailed profiles. Different from them, we directly train a deep personalized character model from a rather larger volume of multimodal conversation data for a single speaker. Detailed comparisons of our multimodal dataset DPCD to the related datasets are presented in Table~\ref{tab:With all other datasets} in Section \ref{subsec: dataset}. 


\begin{table*}[]
  \centering
  \resizebox{\linewidth}{!}{%
  \begin{tabular}{ccccccccccccccccc}
  \toprule[2pt]
  \multirow{2}{*}[-1.5ex]{Dataset} & \multicolumn{3}{c}{Modalities}  & \multicolumn{3}{c}{Extra Attributes}   & \multirow{2}{*}[-1.5ex]{Data Source} & \multirow{2}{*}[-1.5ex]{\# of utterances} & \multirow{2}{*}[-1.5ex]{Total duration} & \multirow{2}{*}[-1.5ex]{\# of speakers} & \multirow{2}{*}{\begin{tabular}[c]{@{}c@{}}Avg.\# of\\ utterances\\ per speaker\end{tabular}} & \multirow{2}{*}{\begin{tabular}[c]{@{}c@{}}Avg.dur.of\\ utterances\\ per speaker\end{tabular}} & \multicolumn{4}{c}{Qualified Tasks}  \\
  \cmidrule(lr){2-4}\cmidrule(lr){5-7}\cmidrule(lr){14-17}
  & V  & A  & T  & \begin{tabular}[c]{@{}c@{}}Conver-\\ sational\end{tabular} & E/S Label  & \begin{tabular}[c]{@{}c@{}}Character\\ Tag\end{tabular} &  & & & & & & \begin{tabular}[c]{@{}c@{}}NLP\end{tabular} & \begin{tabular}[c]{@{}c@{}}AVS \end{tabular} & \begin{tabular}[c]{@{}c@{}}Speech\\ Synthesis\end{tabular} & DPCC   \\ 
  \midrule
  CMU-MOSI~\cite{ex1}    & \checkmark & \checkmark  & \checkmark  &   & \checkmark &  & YouTube & 2199 &2.56h &89 &24.71 &0.029h  & \checkmark  &  &  &   \\
  \specialrule{0em}{1pt}{1.5pt}
  CMU-MOSEI~\cite{ex2}   & \checkmark & \checkmark  & \checkmark  &    & \checkmark &   & YouTube &23453 &65.89h &1000 &23.45 &0.066h  & \checkmark  &  &  &  \\
  \specialrule{0em}{1pt}{1.5pt}
  IEMOCAP~\cite{ex5}     & \checkmark & \checkmark  & \checkmark  & \checkmark    & \checkmark &   & Role Play  &10000 &11.47h &10 &1000 &1.47h & \checkmark   &  &    &   \\
  \specialrule{0em}{1pt}{1.5pt}
  MELD~\cite{ex6}       & \checkmark  & \checkmark  & \checkmark  & \checkmark  & \checkmark &   & TV Series &13708 &12.68h &407 &33.68 &0.031h  & \checkmark  &    &    &   \\
  \specialrule{0em}{1pt}{1.5pt}
  MEISD~\cite{ex7}      & \checkmark  & \checkmark  & \checkmark  & \checkmark  & \checkmark &   & TV Series &20000 &22.22h &4072 &4.91 &5.46e-03h & \checkmark      &    &    &   \\
  \specialrule{0em}{1pt}{1.5pt}
  SEMD~\cite{ex8}       & \checkmark  & \checkmark  & \checkmark   & \checkmark & \checkmark &   & TV Series  &1,066,677 &746h &- &- &- & \checkmark    &  &   &   \\
  \specialrule{0em}{1pt}{1.5pt}
  MSCTD~\cite{ex10}      & \checkmark  &    & \checkmark   & \checkmark   &    &   & Former Dataset  &173,241 &- &- &- &-  & \checkmark   &   &   &    \\
  \specialrule{0em}{1pt}{1.5pt}
  MUStARD~\cite{ex9}    & \checkmark  & \checkmark   & \checkmark   &     & \checkmark & \checkmark  & \begin{tabular}[c]{@{}c@{}}YouTube,MELD \\ and TV Series\end{tabular} &14351 &20.81h &23 &623.96 &0.9h & \checkmark    &   &    &    \\
  \specialrule{0em}{1pt}{1.5pt}
  UR-FUNNY~\cite{ex4}   & \checkmark & \checkmark   & \checkmark   &   & \checkmark &  & TED Talks &63727 &90.23h &1741 &36.6 &0.052h  & \checkmark   &   &  & \\
  \specialrule{0em}{1pt}{1.5pt}
  CH-SIMS~\cite{ex3}    & \checkmark & \checkmark   & \checkmark   &   & \checkmark &  & TV Series &2281 &2.33h &474 &4.81 &4.92e-03h  & \checkmark   &   &  &  \\
  \specialrule{0em}{1pt}{1.5pt}
  CPED\cite{chen2022cped}    & \checkmark & \checkmark   & \checkmark   &\checkmark   & \checkmark &\checkmark  & TV Series &132,762 &78.34h &392 &338.68 &0.20h  & \checkmark   &   &  &  \\
  \specialrule{0em}{1pt}{1.5pt}
  Voxceleb1~\cite{ex11}   & \checkmark & \checkmark   &   &    &   & \checkmark   & YouTube &145,116 &352h &1251 &116 &0.28h  &  & \checkmark  &  &   \\
  \specialrule{0em}{1pt}{1.5pt}
  Voxceleb2~\cite{ex11}   & \checkmark & \checkmark   &   &    &   & \checkmark   & YouTube &\textbf{1,130,720} &2442h &\textbf{6112} &185 &0.40h  &  & \checkmark  &  &   \\
  \specialrule{0em}{1pt}{1.5pt}
  MEAD~\cite{ex12}       & \checkmark & \checkmark   &   &    &   &     & Role Play  &- &38.95h &60 &- &0.65h  &    & \checkmark   &   &  \\
  \specialrule{0em}{1pt}{1.5pt}
  LRW~\cite{ex13}        & \checkmark & \checkmark   & \checkmark   &   &    &   & TV Programs &- &\textbf{1000+h} & 1000+ &- &-  &    & \checkmark    &   &   \\
  \specialrule{0em}{1pt}{1.5pt}
  LJ-Speech~\cite{ex14}  &   & \checkmark   & \checkmark   &    &    & \checkmark   & Book Recording &- &24h &1 &- &-  &    &     & \checkmark  &    \\
  \specialrule{0em}{1pt}{1.5pt}
  CSTR VCTK~\cite{ex15}  &   & \checkmark   & \checkmark   &    &    & \checkmark   & Book Recording &- &44h &110 &- &-  &    &     & \checkmark  &     \\
  \specialrule{0em}{1pt}{1.5pt}
  \textbf{DPCD (Ours)}  & \textbf{\checkmark} & \textbf{\checkmark} & \textbf{\checkmark} & \textbf{\checkmark}  & \textbf{}  & \textbf{\checkmark}   & {TV Series} &48589 &29.65h &5 &\textbf{9717.8} &\textbf{5.93h} & \textbf{\checkmark}   & \textbf{\checkmark}   & \textbf{\checkmark} & \textbf{\checkmark} \\
  \specialrule{0em}{1pt}{1.5pt}
  \bottomrule[2pt]
  \end{tabular}%
  }
  
  \setlength{\belowcaptionskip}{-10pt}
  \caption{A detailed comparison between our dataset DPCD and other multimodal datasets.(E/S Label: Emotion/Sentiment Label)}
  \vspace{-3mm}
  \label{tab:With all other datasets}
  \end{table*}

\section{Deep Personalized Conversation Dataset}


\subsection{Overview of DPCD Collection}
Our goal is to collect character-centric multimodal conversations, from  
 raw videos, audio, transcripts, and subtitles from the TV show. 
 We choose \emph{The Big Bang Theory} as a example in this paper, for its large number of episodes.
The collected items should well reflect the characters' clean contents including facial  expressions, vocals and texts. Meanwhile, this content should be paired and temporally well aligned to allow cross-modality integration. 
To conduct such a large scale data collection and ensure high quality, we adopt an automatic and manual mixed operation as follows: 1) For each modality, we use automatic tools to extract the basic feature, reduce the noise, and roughly align modalities. 2) Manually check the quality of the alignment between text and audio/video modality.
During the automatic processing, for videos, we recognize and track the speaking face, then verify the active speaker and annotate its identity. 
For audio, we use multiple filters and acoustic tool kits to reduce background noise and enhance vocals, and we also remove laughter from the audience to get clean acoustic signals. 
As for the text, we clean and align the transcripts and subtitles through textual similarity computation. 
For feature extraction, we use OpenFace\footnote{https://github.com/cmusatyalab/openface}to extract facial expression features, COVAREP\footnote{https://www.adobe.com/cn/products/premiere.html} to extract acoustic features, and DialoGPT tokenizer\cite{ex31} to get word embeddings.
Finally, we align all three modalities according to the starting and ending timestamps. 
Thanks to the manual efforts, the final dataset includes well-aligned visual, audio, and textual modalities for each utterance, as well as the speaker identity and conversational contexts. Details of the data collection are presented in Supplement Section 1.1 and 1.2. Please note that this collection pipeline can be applied to other TV shows or movies to scale the dataset, and we have verified the effectiveness of the pipeline on \emph{Friends}, see Supplement Section 1.3 for details.

\begin{table}[]
  \centering
  \resizebox{.98\columnwidth}{!}{%
  \begin{tabular}{cccc}
  \toprule[2pt]
  \specialrule{0em}{1pt}{1.5pt}
  \multicolumn{2}{c}{Aligned Utterances (WV)} & \multicolumn{2}{c}{Conversations (W5-R)} \\ 
  \specialrule{0em}{1pt}{1.5pt}
  \midrule
  \# of utterance & 48589  & \# of mono conversations  & 29235  \\
  \specialrule{0em}{1pt}{1.5pt}
  \# of speaker  & 5      & \# of words   & 800356 \\
  \specialrule{0em}{1pt}{1.5pt}
  dura of audios(h)  & 29.65  & Avg. \# of words per turn   & 27.38  \\
  \specialrule{0em}{1pt}{1.5pt}
  Avg. words per utterance          & 7.06   & Avg. dura of mono turn(s)              & 10.17  \\
  \specialrule{0em}{1pt}{1.5pt}
  Avg. \# of utterance per speaker  & 9717.8 & Avg. \# of utterances per mono turn    & 3.67   \\
  \specialrule{0em}{1pt}{1.5pt}
  Avg dura of audio per speaker(h)  & 5.93   & Avg. \# of conversation per speaker    & 5847  \\
  \specialrule{0em}{1pt}{1.5pt}
  \bottomrule[2pt]
  \end{tabular}%
  }
  \setlength{\abovecaptionskip}{0pt} 
  \setlength{\belowcaptionskip}{-10pt}
  \caption{Perfectly aligned utterances with videos (WV) and all conversations with 5 main characters as respondent (W5-R).}
  \label{tab: Dataset stastics}
  \vspace{-1mm}
\end{table}

\subsection{Dataset Exploration}\label{subsec: dataset}


We collect single utterances that are perfectly aligned with corresponding audios to train speech synthesis models. Table~\ref{tab: Dataset stastics} presents several key statistics of the DPCD dataset. We collect 48,589 utterances from 5 characters (Sheldon, Leonard, Penny, Howard, Raj), each utterance contains 7.06 words and lasts 2.20s on average. Since we divide sentences on the basis of subtitles, the average duration for an utterance is relatively short. However, the total duration of audio is 29.65h, reaching a volume qualified for speech synthesis task\cite{ren2019fastspeech,ex32,ex33}.

\begin{figure}
    \centering
    \includegraphics[width=0.8\linewidth]{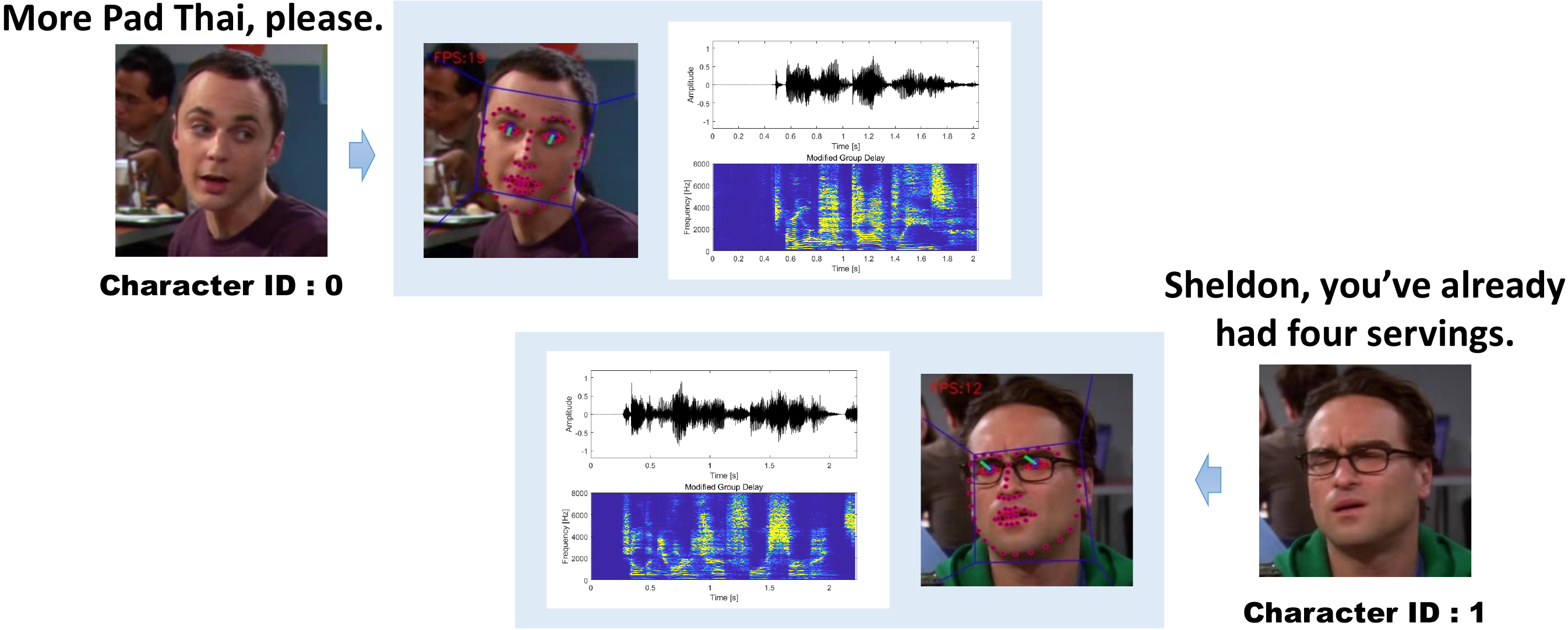}
    \caption{An example from our dataset. A sentence in a dialogue is aligned with corresponding audio, video, speaker identity and extracted visual/acoustic features.}
    \label{fig:short-b}
  \label{fig:Dataexample}
\end{figure}

We also maintain complete conversational order of the utterances, and extract mono conversation turns as an example. Using these conversation data, we can learn character's interaction pattern and speaking style, and try to model that character's personality. Figure~\ref{fig:Dataexample} gives a brief example of conversation data in our DPCD dataset. Our DPCD dataset collects 29,235 mono conversation turns in total. Each conversation turn contains about 27.38 words and lasts 10.17s. As conversations involve interaction between speakers, the corpus has a richer collection of utterances and is more expressive.

Detailed comparison of DPCD to the related datasets is presented in Table~\ref{tab:With all other datasets}. According to the targeting tasks, we roughly divide the multimodal datasets into three categories for natural language processing (NLP), audio-visual synchronization (AVS) and speech synthesis, respectively. The existing multimodal datasets for NLP \cite{ex1,ex2,ex3,ex4,ex5,ex6,ex7,ex8,ex9,ex10} are usually designed to understand interaction patterns or detect the fine-grained conversation features in dialogue turns. However, they are typically collected from a wide range of speakers to balance gender and race, resulting in inadequate data per speaker for tasks like personalized character encoding. Audio-visual datasets \cite{ex11,ex12,ex13} for AVS typically consist of huge amounts of high-quality talking head videos. However, since it's rather challenging to extract subtitles from in-the-wild videos, these datasets hardly contain aligned texts or conversation contexts, and thus are not qualified for most conversation tasks. Datasets for speech synthesis \cite{ex14,ex15} usually contain elaborately aligned text and acoustic modalities. They are always recorded in a noiseless studio to guarantee acoustic quality. Transcripts are forced aligned with audio clips to maintain accurate correspondence relationship between phonemes and pronunciation.  All of the existing multimodal datasets are always collected to qualify a specific task in a single field, hence leaving some modalities of poor quality. With the total scale of data increasing, the data size for each character is often ignored. 

Different from most multimodal datasets targeting NLP tasks, the videos and audios offered by our DPCD dataset are more informative and clean. Our videos are specifically cropped to focus on active talking face, excluding other characters and complicated background scenes. So the visual features extracted from these video crops can better represent facial expressions and the speaker's emotions, free from possible environmental noises. The audios in our dataset are also elaborately filtered to better capture the speaker's voice and tone. Compared with many audio-visual datasets, our texts are labeled with speaker identity and conversational context, available for further expression and emotion modeling.
In this way, our dataset DPCD is qualified for tasks in the fields such as natural language processing, audio-visual synchronization and speech synthesis.

\begin{table}[]
  \centering
  \resizebox{.98\columnwidth}{!}{%
  \begin{tabular}{cccccc}
  \toprule[2pt]
  Detailed Statistics  & Sheldon  & Penny  & Leonard  & Howard  & Raj   \\ 
  \midrule
  \# of utterance    & 18747  & 7636  & 9803  & 6557 & 5846   \\
  \specialrule{0em}{1pt}{1.5pt}
  dura of audios(h)  & 12.10  & 5.68  & 4.30  & 4.07 & 3.49    \\
  \specialrule{0em}{1pt}{1.5pt}
  \begin{tabular}[c]{@{}c@{}}Avg. words per utterance\end{tabular} & 7.05 & 7.09  & 7.05  & 7.05 & 7.13  \\
  \specialrule{0em}{1pt}{1.5pt}
  \# of conversation turns  & 8564  & 5566  & 7075  & 4434  & 3633  \\
  \specialrule{0em}{1pt}{1.5pt}
  \begin{tabular}[c]{@{}c@{}}Avg. words  per response\end{tabular} & 17.32 & 12.05 & 12.01 & 13.01 & 13.84 \\ 
  \bottomrule[2pt]
  \end{tabular}%
  }
  \setlength{\belowcaptionskip}{-10pt}
  \caption{Detailed statistics on characters in DPCD.}
  \label{tab: Dataset stastics per character}
\end{table}

\begin{figure*}[t]
  \centering
  \includegraphics[width=0.98\linewidth]{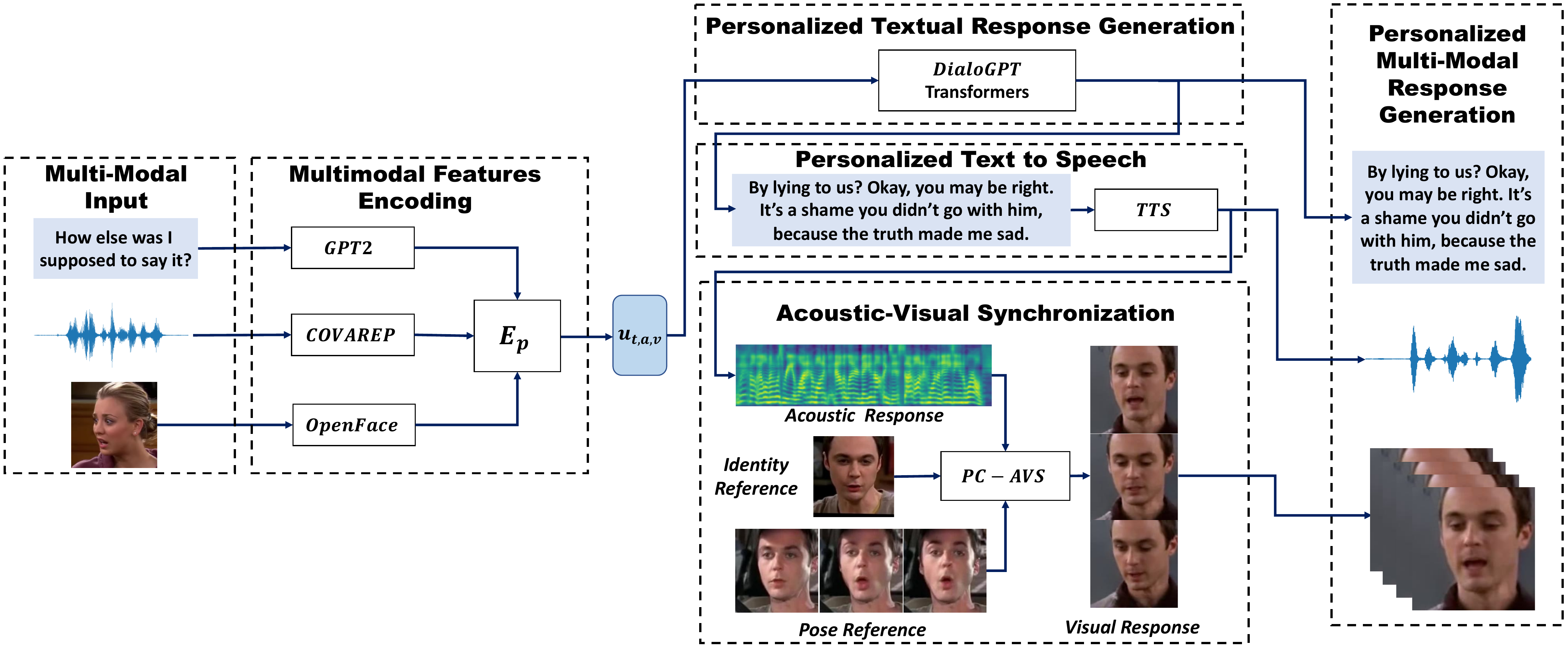}
   \caption{Overview of our baseline solution for DPCC. Given multimodal input, we first extract multimodal features, 
   then the personalized textual, speech and video responses are generated by a personalized textual response generation model, a personalized text-to-speech model, and an acoustic-visual synchronization model, sequentially and respectively.}
   \label{fig:Task}
\end{figure*}

Furthermore, as mentioned above, our DPCD dataset mainly focuses on 5 characters, detailed statistics for each character are presented in Table~\ref{tab: Dataset stastics per character}. It is obvious, from the table, that our dataset offers a competitively large volume of data for a single character. Sheldon acted as a respondent in 8,564 mono conversation turns, each containing about 17.32 words. While Penny participates in 5,566 conversation turns, each containing about 12.05 words. The average utterance audio duration per character reaches 5.93h, average utterance number per character reaches 9,717.8, greatly exceeding all the existing multimodal datasets. 
This relatively large volume of character data makes it possible to model personality through multimodal inputs. To be specific, as we offer sufficient utterance data and conversational data, it is possible to model the character's speaking manner and interaction habits. With hours of audio tracks, one can capture the character's voice and tone to build a personalized speech synthesis model. One can also rebuild the character's visual image or 3D model from the video clips, and analyze the character's facial expression tendency during the conversation.


\section{Our DPCC Baseline}

In this section, we provide a baseline for DPCC, and the whole pipeline is presented in Figure~\ref{fig:Task}. 

\textbf{Multimodal Features Encoding.} 
We train a multimodal feature representation model to capture each modality's distinctive stylistic information. Following MISA\cite{ex30}, one-dimensional semantic features are previously extracted from each utterance or audio/video clip, then projected into a modality-specific space utilizing feature projection model $E_p$. $E_p$ is pretrained on a multimodal emotion recognition task\cite{ex2,zadeh2018multi} for better understanding in multimodal correlation. The textual/acoustic/visual features are encoded by $E_p$ to form multimodal conversation context input $u_{t,a,v}$.

\textbf{Personalized Textual Response Generation.} We finetune the pretrained model conversational response generation DialoGPT\cite{ex31} on a single character's multimodal conversation context input $u_{t,a,v}$ to model his personalized interaction pattern and speaking style and finally generate the specific character's possible response to the multimodal input context.

\textbf{Personalized Text to Speech.} 
We use the TTS model proposed by Jia et al.\cite{jia2018transfer} to utter the generated textual response in the target character's voice and tone. For each character, the TTS model is finetuned on wave files and utterances collected from that specific character to capture unique speech patterns.



\textbf{Acoustic-Visual Synchronization.} Utilizing the acoustic-visual synchronization model, we synthesize our generated textual response and corresponding audio to the final multimodal response. We adopt the pretrained Pose-Controllable Talking Face Generation Model(PC-AVS) proposed by Zhou et al\cite{ex35}. The identity reference and pose reference are randomly selected from the target character's video clips and lip movements are synchronized with the wave file generated by the personalized speech synthesis model. In this way, all generated modalities are aligned and synthesized to a final multimodal personalized response.

\section{Experiments}

\subsection{Experimental setup}

\textbf{Baseline.} We use the multimodal data collected in our dataset to train the baseline DeepCharacter, and separate models are trained for different characters, respectively. To pretrain multimodal feature representation models, we utilize multimodal features and emotion labels from CMU-MOSEI dataset\cite{ex2}. We set the batch size to 8 and hidden feature size to 64, and use the original default settings. The encoded 64-dimension audio/vision features are concatenated with 128-dimension word embeddings transformed by pretrained GPT2 tokenizer. 
Then we use this multimodal input to finetune DialoGPT-small conversation response generation model.
See Supplement Section 3 for detail setting. 
As for the personalized speech synthesis model, the audios are transformed to 22050Hz sample rate first and aligned with utterances using Montreal Forced Aligner(MFA)\footnote{https://github.com/MontrealCorpusTools/Montreal-Forced-Aligner}. Videos should be transformed to 25fps and wavfiles to 16kHz for Pose Controllable Talking Face Generation, and all other settings by default. All models are implemented by PyTorch\cite{ex36} and performed on a RTX-3090 GPU.

\textbf{Evaluation.}
Since our proposed task and baseline method heads towards creating personalized characters, the degree of characterization is a key benchmark to evaluate the model performance. We evaluate both the characterization and quality. In spite of objective experiments, we also conduct subjective experiment to evaluate the user experience. The objective evaluation is on all the 5 DeepCharacters: DeepSheldon, DeepPenny, DeepLeonard, DeepHoward and DeepRaj. 
For the subjective evaluation, we conduct a survey on 50 people and choose 30 judges according to their knowledge of the characters, see Supplement Section 2.1 for the standard details. The survey shows that three characters Sheldon, Leonard and Penny, are more impressed and related conclusion should be more reliable, thus we conduct subjective evaluation on the corresponding 3 DeepCharacters: DeepSheldon, DeepPenny and DeepLeonard.
Since the textual response lays the foundation of multimodal response, and psychologists think that language can express inner thoughts and feelings\cite{AhmadAsgharKhanHabib2020a}, so it's possible to distinguish a specific character according to his/her utterances\cite{Wang2020encoding}. Therefore, we conduct experiments on both textual and multimodel ones.


\subsection{Characterization Evaluation}

\begin{table}[]
  \centering
  \resizebox{.98\columnwidth}{!}{%
  \begin{tabular}{m{7cm}m{4.5cm}<{\centering}m{4.5cm}<{\centering}m{4.5cm}<{\centering}} 
  \toprule[2pt]
  & DeepSheldon    & DeepPenny    & DeepLeonard   \\
  \midrule[1.5pt] 
  C1: Dammit man, we’re dying here.    & No we aren’t. This is a swirling vortex of entropy.  & Yeah, I know.   & What are you doing? \\
  \specialrule{0em}{1.5pt}{1.5pt}
  \hline
  \specialrule{0em}{1.5pt}{1.5pt}
  C2: You know, if you guys are interested, there’s a technique where I can take a lock of your hair, refine it into carbon dust and use the hydraulic press at work to turn it into a tiny little diamond with your DNA in it. & I’m sorry, but that’d be impossible. The reaction time would be too much. & Oh my God, that is so stupid. & That sounds fun.\\
  \bottomrule[2pt]
  \end{tabular}%
  }
  \caption{The response of different characters to same textual input.}
  \label{tab:Q and A}
\end{table}

\textbf{Characterization of textual responses.} 
As a vivid illustration, we present 2 samples of different characters response to same textual inputs in Table~\ref{tab:Q and A}. Character's personality can be somehow reflected in these different responses: Sheldon is a high IQ scientist who uses scientific terms like 'vortex of entropy' or 'reaction time' from time to time; Penny, on the other hand, is a woman with rich affection, preferring interjection like 'Yeah','Oh god' or 'Sweetie'; Leonard, although also being somewhat bookish, is more like a normal and helpful person who questions strange utterances and tend to give favorable replies. And as proved by characterization clustering, character classification and human evaluation, our DeepCharacter models can generate well-characterized textual responses.

Following a previous work~\cite{vishnubhotla-etal-2019-fictional} on character classification, we use SAGE\cite{eisenstein2011sparse} model to derive weights for words uttered by the characters.  SAGE enforces a sparse prior on its parameters, and may be sensitive to infrequent terms in the text. 
To alleviate this issue, given one character, we experimentally group 100 randomly selected responses from this character, forming a virtual document\cite{VM2022-TED}, and reweight the words inside the document utilizing SAGE. 
Then we use a pretrained basic BERT model to encode the reweighted virtual documents into 128-dimensional embeddings as the final representation of one sample of the given character's response. 
We perform t-SNE\cite{JMLR:v9:vandermaaten08a} on these sampled representations from all the 5 charaters and 5 DeepCharacters. 
Note that during the above process, BERT are fixed. See details in Supplement Section 3.3 for implementation details. 

\begin{figure}[t]
  \centering
  \includegraphics[width=0.98\linewidth]{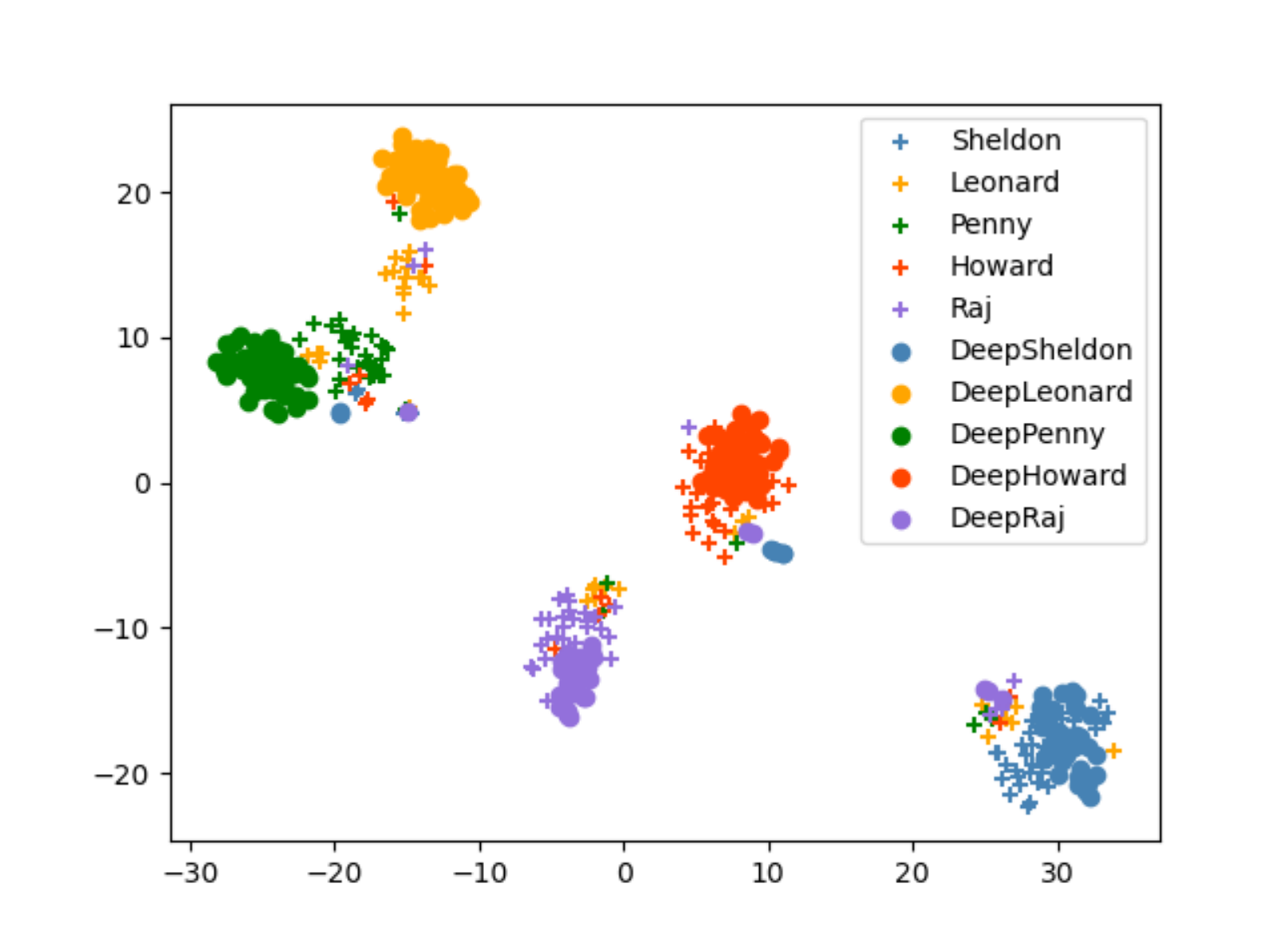}

   \caption{t-SNE visualization of characters' original responses and DeepCharacters' generated ones. Different colour represents different characters. Cross and circle mark represents original and generated text, respectively. }
   \label{fig:CharacterClassify}
\end{figure}

As presented in Figure~\ref{fig:CharacterClassify}, the responses from 5 characters can be well separately clustered, showing the character's diversity in terms of speaking style and word preference. 
Intriguingly, the clusters of responses DeepCharacters aligns quite well with the ones of the corresponding characters, respectively. This verifies that the created DeepCharacters can generate highly personalized textual responses, well capturing the speaking style and word preference.

To provide quantative results, we also build a character classifier based on BERT to automatically evaluate the characterization of text responses generated by our DeepCharacters.  
The character classifier is trained on 18,000 documents from original corpus.  See details in Supplement Section 3.3 for implementation details.  
This character classifier achieves 98.3\% validation accuracy on origin corpus (900 documents) while getting 86.4\% testing accuracy on generated responses (2,000 documents), showing that our DeepCharacters can well capture the characteristics. More discussions on utterance grouping number and more visualization results are provided in Supplement Section 4.2.

For human evaluation, we randomly select 50 response samples from {DeepSheldon}, {DeepPenny} and {DeepLeonard} respectively, and each is paired with general response generated by DialoGPT as a non-personalized baseline, and response generated by other character model as an untargeted character comparison. 
Each question is presented to at least 3 judges. 

The judges are asked to rank each response pair for how well the response matches the given material in speaking style, decision-making style and emotion tendency, using a 3-point Likert-like scale following~\cite{ex31}. 
As shown in~Table~\ref{tab:Human evaluation results1}, all three targeted DeepCharacters gain more preferences over non-personalized DialoGPT and untargeted DeepCharacters, which shows our DeepCharacter's ability of capturing a specific character's personality and speaking style. 

\begin{table}[t]
  \centering
  \setlength{\tabcolsep}{5mm}
  \resizebox{.98\columnwidth}{!}{%
  \begin{tabular}{cccc}
  \toprule[2pt]
  Model                               & Sheldon                    & Penny                      & Leonard                   \\
  \midrule[1.5pt]
  \specialrule{0em}{1pt}{1.5pt}
  DialoGPT                          & 0.420                      & 0.058                      & 0.439                     \\
  Neutral                             & 0.077                      & 0.321                      & 0.058                     \\
  DeepCharacter                      & \textbf{0.503***}             & \textbf{0.622***}             & \textbf{0.503***}            \\
  \hline
  \specialrule{0em}{1pt}{1pt}
  Other DeepCharacters             & 0.367                      & 0.342                      & 0.465                     \\
  Neutral                             & 0.042                      & 0.065                      & 0.026                     \\
  DeepCharacter                      & \textbf{0.590***}             & \textbf{0.594***}             & \textbf{0.510***}            \\
  \bottomrule[2pt]
  \end{tabular}%
  }
  \caption{Results of Human Evaluation for characterization in textual responses, showing preferences for our model: DeepCharacter vis-a-vis DialoGPT or other DeepCharacter model. Numbers in bold indicate the preferred models. Here, \emph{other DeepCharacters} refers to DeepCharacter trained on characters other than the target one, and \emph{Neutral} refers to neutral option without preference for either model.  Statistically significant responses are indicated: * p$\le$0.05, ** p$\le$0.01, *** p$\le$0.001.}
  \label{tab:Human evaluation results1}
\end{table}

\begin{table}[t]
  \centering
  \resizebox{.98\columnwidth}{!}{%
  \begin{tabular}{ccccc}
  \toprule[2pt]
  \multirow{2}{*}{Multimodal Responses}   &\multirow{2}{*}{Model}     & \multirow{2}{*}{Sheldon} & \multirow{2}{*}{Penny} & \multirow{2}{*}{Leonard}\\
  &    &       &     & \\
  \midrule[1.5pt]
  \specialrule{0em}{1pt}{1pt}
  \multirow{2}{*}{Multimodal Quality}
  & Baseline       & 0.29                      & 0.33                      & 0.29                     \\
  & DeepCharacter   & \textbf{0.71***}             & \textbf{0.67***}             & \textbf{0.71***} \\
  \hline
  \specialrule{0em}{1pt}{1pt}
  \multirow{2}{*}{Characterization}
  & Baseline   & 0.10                      & 0.32                      & 0.16\\
  & DeepCharacter & \textbf{0.90***}             & \textbf{0.68***}             & \textbf{0.84***}   \\
  \bottomrule[2pt]
  \end{tabular}%
  }
  \caption{Results of Human Evaluation for multimodal responses, showing preferences for our model: DeepCharacter vis-a-vis neutral baseline. Numbers in bold indicate the preferred models.Statistically significant responses are indicated: *** p$\le$0.001. } 
  \label{tab:User study results}
\end{table}

\textbf{Characterization of multimodal responses.}
As characterization in multi-modality is comparatively subjective, we rely on user study to evaluate our DeepCharacter's capacity in generalizing personalized multimodal responses. 
We use DialoGPT finetuned on whole DPCD corpus instead of any specific character, and un-finetuned TTS model as the neutral baseline, while the inference of acoustic-visual synchronization model remains the same as DeepCharacter. 
The neutral baseline has learned the TV show corpus's special text distribution and can generate high quality speeches.

We randomly select 10 video clips generated by DeepSheldon, DeepLeonard and DeepPenny, respectively. 
Like PC-AVS\cite{ex35}, the total 30 video clips, paired with baseline results, are handed to all 30 judges. 
The judges are asked to rank each multimodal response pair base on how well the response matches the target character in terms of speaking style, tone and identity, considering all three modalities comprehensively. 
As shown in Table~\ref{tab:User study results} , all three targeted DeepCharacters gain more preferences over the neutral baseline, which shows our DeepCharacter's ability of generating characterized multimodal responses.

The above results demonstrate the collected DPCD can support our simple baseline to generate deep personalized characters.

\begin{table}[]
  \centering
  \resizebox{\columnwidth}{!}{%
  \begin{tabular}{ccccccccccccc}
  \toprule[2pt]
  \multirow{2}{*}{Model} & \multicolumn{2}{c}{Dist-1} & \multicolumn{2}{c}{Dist-2} & \multicolumn{2}{c}{Avg.len} & \multicolumn{2}{c}{Entropy-4} & \multicolumn{2}{c}{Perplexity} & \multirow{2}{*}{SacreBleu} \\
  \cmidrule(lr){2-3}\cmidrule(lr){4-5}\cmidrule(lr){6-7}\cmidrule(lr){8-9}\cmidrule(lr){10-11}
  \specialrule{0em}{1pt}{1.5pt}
  & O-M        & R-D         & O-M         & R-D         & O-M          & R-D          & O-M           & R-D           & O-M            & R-D           &                            \\ 
  \specialrule{0em}{1pt}{1.5pt}
  \midrule
  DeepSheldon                    & 0.413       & 0.465       & 0.841       & 0.902       & 14.01        & 14.90        & 7.68          & 7.76          & 172.687        & 174.365            & 0.510 \\
  \specialrule{0em}{1pt}{1.5pt}
  DeepPenny                      & 0.341       & 0.401       & 0.798       & 0.882       & 10.18        & 12.37        & 7.23          & 7.53          & 131.339        & 144.014            & 1.850 \\
  \specialrule{0em}{1pt}{1.5pt}
  DeepLeonard                    & 0.383       & 0.422       & 0.813       & 0.892       & 9.56         & 10.32        & 7.16          & 7.28          & 142.605        & 167.472       & 1.183 \\
  \specialrule{0em}{1pt}{1.5pt}
  DeepHoward                    & 0.422       & 0.442       & 0.855       & 0.897       & 13.42        & 14.08        & 7.512          & 7.597          & 176.661        & 177.023         & 0.503 \\
  \specialrule{0em}{1pt}{1.5pt}
  DeepRaj                       & 0.423       & 0.448       & 0.874       & 0.917       & 13.66        & 14.21        & 7.511          & 7.616          & 241.974        & 215.250         & 0.345 \\
  \specialrule{0em}{1pt}{1.5pt}
  DialoGPT                   & \multicolumn{2}{c}{0.217} & \multicolumn{2}{c}{0.414} & \multicolumn{2}{c}{7.955} & \multicolumn{2}{c}{6.33} & \multicolumn{2}{c}{34.561} & 0.160 \\
  DialoGPT (finetuned)       & \multicolumn{2}{c}{0.384} & \multicolumn{2}{c}{0.794} & \multicolumn{2}{c}{11.475} & \multicolumn{2}{c}{7.25} & \multicolumn{2}{c}{163.76} & 0.254 \\
  \specialrule{0em}{1pt}{1.5pt}
  \bottomrule[2pt]
  \end{tabular}%
  }
  \caption{Automatic evaluation scores including O-M (Our Method) and R-D (Reference Data).}
  \label{tab:Automatic evaluation results}
  \end{table}
  
\subsection{Quality evaluation}
We perform automatic and human evaluations on the quality of responses generated by our DeepCharacter model. Here, high quality responses refer to  reasonable texts, realistic speeches and videos, and of high naturalness and consistency across all three modalities.

\textbf{Quality of textual responses.}
Following DialoGPT\cite{ex31}, we perform automatic evaluations using several popular standard evaluation metrics, including ScareBLEU\cite{ex39} and Perplexity\cite{ex38}. SacreBLEU provides hassle-free computation of BLEU scores, ranging from 0 to 100, and a higher SacreBLEU means a better match between generated results and references. Perplexity measures how likely the model is to generate the input text sequence and can be used to evaluate how well the model has learned the distribution of the training text.  We also use Entropy\cite{ex41} and Dist-n\cite{ex40} to evaluate lexical diversity.

We calculate automatic evaluation scores for DeepSheldon, DeepPenny, DeepLeonard, DeepHoward and DeepRaj separately. In comparison, we evaluate DialoGPT's results with the same contexts, as a non-personalized general response. We also finetune DialoGPT on the mixed corpus of these characters, to bridge the performance gap caused by corpus difference and serve as an un-targeted response. We additionally evaluate the lexical diversity and perplexity of the characters' original corpus, to learn text distribution.

Table~\ref{tab:Automatic evaluation results} summarizes the automatic evaluation results, showing DeepCharacter's stronger adaptability to complicated daily problems. All DeepCharacters achieve a higher score in lexical diversity than original DialoGPT, and are close to respective original corpus. Both generated responses and original corpus have higher perplexity than DialoGPT's results, illustrating the complexity and distribution diversity of our DPCD, which can also be verified by the results of finetuned DialoGPT. 
This also demonstrates DeepCharacter's ability to fit text distribution of the original corpus. 
Due to high complexity and ambiguity of generation task, all five DeepCharacters receive rather low SacreBLEU score, but still higher than original and finetuned DialoGPT, showing a stronger reasoning ability to understand and reply diverse and colloquial dialogues. 

\begin{table}[]
  \centering
  \resizebox{.98\columnwidth}{!}{%
  \begin{tabular}{cccccc}
  \toprule[2pt]
  \multirow{2}{*}{Elements}   &\multirow{2}{*}{Model}     & \multirow{2}{*}{Sheldon} & \multirow{2}{*}{Penny} & \multirow{2}{*}{Leonard} & \multirow{2}{*}{Avg.of  Model} \\
  &    &       &     &     &     \\
  \midrule[1.5pt]
  \specialrule{0em}{1pt}{1pt}
  \multirow{2}{*}{Fluency}
  & DialoGPT     & -         & -    & -      & 2.646                                \\
  & \textbf{DeepCharacter} & 2.720   & 2.602   & 2.630    & \textbf{2.651$^+$}      \\
  \hline
  \specialrule{0em}{1pt}{1pt}
  \multirow{2}{*}{Context Relevance}
  & \textbf{DialoGPT}    & -         & -    & -      & \textbf{2.169$^+$}       \\
  & DeepCharacter & 2.001   & 2.240   & 2.216    & 2.152       \\
  \hline
  \specialrule{0em}{1pt}{1pt}
  \multirow{2}{*}{Emotional Degree}
  & DialoGPT    & -         & -    & -      & 2.066                                    \\
  & \textbf{DeepCharacter} & 2.048  & 2.161    & 2.090   & \textbf{2.100$^+$}                       \\ 
  \bottomrule[2pt]
  \end{tabular}%
  }
  \caption{Results of Human Evaluation for text quality, showing preferences for our model: DeepCharacter vis-a-vis DialoGPT. The average text quality score of our three DeepCharacters is calculated and compared with other models in the last column, and numbers in bold indicate the preferred models. Statistically significant responses are indicated: $^+$ p$\ge$0.5} 
  \label{tab:Human evaluation results2}
\end{table}

We also conduct human study to evaluate our model's ability in generating high quality characterized textual responses. The testing data remain same with characterization evaluation.
We first evaluate the generated text quality from the views of fluency, context relevant and personalized/emotional degree. Judges rate between 1 to 3 for each aspect, where 1 means "very bad, not at all fluent/not at all relevant/very general", 2 means "not so bad, some minor mistakes", and 3 means "very fluent/relevant/personalized or emotional". 

As shown in Table~\ref{tab:Human evaluation results2}, all DeepCharacters reach high scores on all three evaluation criteria. 
According to significance test, the differences between DeepCharacter and DialoGPT in these three criteria are less significant, indicating that our DeepCharacter can generate responses that are as fluent, reasonable, and emotional as DialoGPT. 
While DialoGPT is well-known for generating high quality responses given basic and general contexts, our DeepCharacter can achieve similar quality in complicated personalized scenarios.


The above results demonstrate that our DeepCharacters achieve comparatively high score in lexical diversity, reasonalibity, fluency and emotionality of generated texts. 

\textbf{Quality of generated multimodal responses.}
We conduct user study to evaluate the multimodal quality of the generated videos. The data and experiment settings remain the same with the video characterization user study, and the judges are required to rank each multimodal response pair for the responses' performance in language fluency, naturalness and realness. 

As shown in Table~\ref{tab:User study results}, all three DeepCharacters gain more preferences over comparative baseline, which confirms our DeepCharacter's ability of generating high quality multimodal responses. 
Although videos generated by both models shares similar speech clarity and image sharpness, the vivid tone and reasonable textual responses generated by our DeepCharacter increase naturalness and realness, resulting in judges' preference when considering multimodal quality. Please see the Supplementary Materials for the multimodal survey contents.

\begin{figure}[t]
  \centering
  \includegraphics[width=0.98\linewidth]{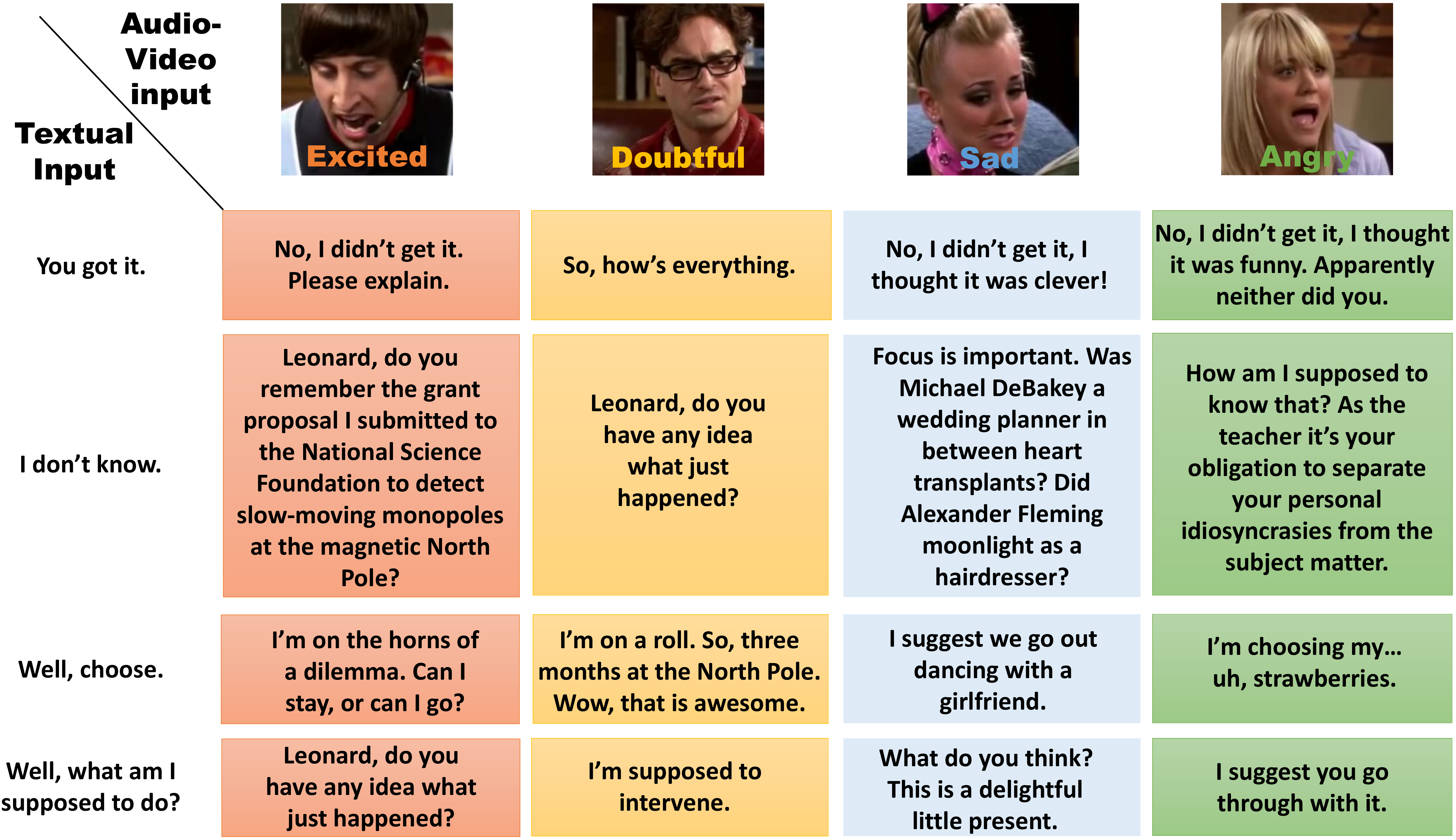}

   \caption{Examples of DeepSheldon response with multimodal inputs. For each row, given the inputs of the same textual context and but different video/audio context with different emotions (labelled with different colors). 
   }
   \label{fig:GenerationExample}
\end{figure}

To study how the multimodal input affects the response, we take the same textual stimuli mixed with different video/audio context. As shown in Figure~\ref{fig:GenerationExample}, instead of generating the same response to the shared text, our model gives different results not only in response to the textual context, but also related to the emotion expressed in the video/audio context. 
Specifically, our DeepCharacter tends to generate a more tender or more positive response when given video/audio inputs expressing sad moods. On the contrary, when the speaker in the video/audio input is very angry and unfriendly, the model will be more likely to generate unkind words, sometimes seeming to fight back, like "\emph{How am I supposed to know that?}", or "\emph{No, I didn’t get it, I thought it was funny. Apparently neither did you.}" Extra modalities can offer more effective information, such as the speaker's facial expressions, voice or tone, which can better express the speaker's emotion and will certainly affect natural response generation. Given multimodal inputs, the model can obtain more comprehensive information from the conversation context, thus understanding the character's interaction pattern and personality better, generating diverse and more personalized responses.

\section{Conclusion}
Towards a fantastic goal of creating personalized characters with whom one can interact in a multimodal channel, we explore creating personalized characters from collected multimodal data, in a data-driven way. In this paper, we formulate a novel task (DPCC), given multimodal-in stimuli, and predict multimodal-out response. 
To support this task, we collect a multimodal conversation dataset (DPCD) surrounding several characters from TV shows. This dataset provides about 10 times more multimodal conversations of high quality, well aligned across multimodality. 
We further provide a baseline solution for learning multimodal response given multimodal input. 
The created DeepCharacters is capable of generating relatively high-quality response, and consistent personality to the original character. 
In the future, we would like to collect richer characters' data such as pose, gestures etc, and more rounds of conversations to further enrich the personalized feature. 
We suppose this work can shed light on an interesting research direction towards creating deep personalized characters to support multimodal chatting scenarios, and inspire further works on, such as, 
improving the visual and acoustic quality, proposing metrics for evaluating personality similarity and fidelity, extending the 2D video to 3D for more immersive interaction.

{\small
\bibliographystyle{ieee_fullname}
\bibliography{egbib}

\begin{thebibliography}{10}\itemsep=-1pt

\bibitem{AhmadAsgharKhanHabib2020a}
Hussain Ahmad, Muhammad~Zubair Asghar, Alam~Sher Khan, and Anam Habib.
\newblock A systematic literature review of personality trait classification
  from textual content.
\newblock {\em Open Computer Science}, 10(1):175--193, 2020.

\bibitem{arik2017deep}
Sercan~{\"O} Ar{\i}k, Mike Chrzanowski, Adam Coates, Gregory Diamos, Andrew
  Gibiansky, Yongguo Kang, Xian Li, John Miller, Andrew Ng, Jonathan Raiman,
  et~al.
\newblock Deep voice: Real-time neural text-to-speech.
\newblock In {\em International Conference on Machine Learning (ICML)}, pages
  195--204. PMLR, 2017.

\bibitem{ex2}
AmirAli Bagher~Zadeh, Paul~Pu Liang, Soujanya Poria, Erik Cambria, and
  Louis-Philippe Morency.
\newblock Multimodal language analysis in the wild: {CMU}-{MOSEI} dataset and
  interpretable dynamic fusion graph.
\newblock In {\em Proceedings of the 56th Annual Meeting of the Association for
  Computational Linguistics (ACL)}, pages 2236--2246, Melbourne, Australia,
  July 2018. Association for Computational Linguistics.

\bibitem{ex5}
Carlos Busso, Murtaza Bulut, Chi-Chun Lee, Abe Kazemzadeh, Emily Mower,
  Jeannette~N. Chang, Samuel Kim, Sungbok Lee, and Shrikanth~S. Narayanan.
\newblock {IEMOCAP}: interactive emotional dyadic motion capture database.
\newblock In {\em Language Resources and Evaluation}, pages 335--359, Dec.
  2008.

\bibitem{ex9}
Santiago Castro, Devamanyu Hazarika, Ver{\'o}nica P{\'e}rez-Rosas, Roger
  Zimmermann, Rada Mihalcea, and Soujanya Poria.
\newblock Towards multimodal sarcasm detection (an {\_}{O}bviously{\_} perfect
  paper).
\newblock In {\em Proceedings of the 57th Annual Meeting of the Association for
  Computational Linguistics (ACL)}, pages 4619--4629, Florence, Italy, July
  2019. Association for Computational Linguistics.

\bibitem{ex33}
Mingjian Chen, Xu Tan, Bohan Li, Yanqing Liu, Tao Qin, Sheng Zhao, and Tie-Yan
  Liu.
\newblock Adaspeech: Adaptive text to speech for custom voice, 2021.

\bibitem{chen2022cped}
Yirong Chen, Weiquan Fan, Xiaofen Xing, Jianxin Pang, Minlie Huang, Wenjing
  Han, Qianfeng Tie, and Xiangmin Xu.
\newblock {CPED}: A large-scale chinese personalized and emotional dialogue
  dataset for conversational ai.
\newblock {\em arXiv preprint arXiv:2205.14727}, 2022.

\bibitem{ex13}
Joon~Son Chung and Andrew Zisserman.
\newblock Lip reading in the wild.
\newblock In Shang-Hong Lai, Vincent Lepetit, Ko Nishino, and Yoichi Sato,
  editors, {\em Computer Vision -- ACCV 2016}, pages 87--103, Cham, 2017.
  Springer International Publishing.

\bibitem{eisenstein2011sparse}
Jacob Eisenstein, Amr Ahmed, and Eric~P. Xing.
\newblock Sparse additive generative models of text.
\newblock In {\em Proceedings of the 28th International Conference on
  International Conference on Machine Learning}, ICML'11, page 1041–1048,
  Madison, WI, USA, 2011. Omnipress.

\bibitem{ex7}
Mauajama Firdaus, Hardik Chauhan, Asif Ekbal, and Pushpak Bhattacharyya.
\newblock {MEISD}: A multimodal multi-label emotion, intensity and sentiment
  dialogue dataset for emotion recognition and sentiment analysis in
  conversations.
\newblock In {\em Proceedings of the 28th International Conference on
  Computational Linguistics (COLING)}, pages 4441--4453, Barcelona, Spain
  (Online), Dec. 2020. International Committee on Computational Linguistics.

\bibitem{ex8}
Mauajama Firdaus, Hardik Chauhan, Asif Ekbal, and Pushpak Bhattacharyya.
\newblock Emosen: Generating sentiment and emotion controlled responses in a
  multimodal dialogue system.
\newblock {\em IEEE Transactions on Affective Computing}, 13(3):1555--1566,
  2022.

\bibitem{ex4}
Md~Kamrul Hasan, Wasifur Rahman, AmirAli Bagher~Zadeh, Jianyuan Zhong,
  Md~Iftekhar Tanveer, Louis-Philippe Morency, and Mohammed~(Ehsan) Hoque.
\newblock {UR}-{FUNNY}: A multimodal language dataset for understanding humor.
\newblock In {\em Proceedings of the 2019 Conference on Empirical Methods in
  Natural Language Processing and the 9th International Joint Conference on
  Natural Language Processing (EMNLP-IJCNLP)}, pages 2046--2056, Hong Kong,
  China, Nov. 2019. Association for Computational Linguistics.

\bibitem{ex30}
Devamanyu Hazarika, Roger Zimmermann, and Soujanya Poria.
\newblock Misa: Modality-invariant and -specific representations for multimodal
  sentiment analysis.
\newblock In {\em Proceedings of the 28th ACM International Conference on
  Multimedia (ACMMM)}, MM '20, page 1122–1131, New York, NY, USA, 2020.
  Association for Computing Machinery.

\bibitem{ex14}
Keith Ito and Linda Johnson.
\newblock The lj speech dataset.
\newblock \url{https://keithito.com/LJ-Speech-Dataset/}, 2017.

\bibitem{ex38}
Fred Jelinek, Robert~L Mercer, Lalit~R Bahl, and James~K Baker.
\newblock Perplexity—a measure of the difficulty of speech recognition tasks.
\newblock {\em The Journal of the Acoustical Society of America},
  62(S1):S63--S63, 1977.

\bibitem{jia2018transfer}
Ye Jia, Yu Zhang, Ron~J. Weiss, Quan Wang, Jonathan Shen, Fei Ren, Zhifeng
  Chen, Patrick Nguyen, Ruoming Pang, Ignacio~Lopez Moreno, and Yonghui Wu.
\newblock Transfer learning from speaker verification to multispeaker
  text-to-speech synthesis.
\newblock In {\em Proceedings of the 32nd International Conference on Neural
  Information Processing Systems}, NIPS'18, page 4485–4495, Red Hook, NY,
  USA, 2018. Curran Associates Inc.

\bibitem{ex40}
Jiwei Li, Michel Galley, Chris Brockett, Jianfeng Gao, and Bill Dolan.
\newblock A diversity-promoting objective function for neural conversation
  models.
\newblock In {\em Proceedings of the 2016 Conference of the North {A}merican
  Chapter of the Association for Computational Linguistics: Human Language
  Technologies (NAACL-HLT)}, pages 110--119, San Diego, California, June 2016.
  Association for Computational Linguistics.

\bibitem{ex16}
Jiwei Li, Michel Galley, Chris Brockett, Georgios Spithourakis, Jianfeng Gao,
  and Bill Dolan.
\newblock A persona-based neural conversation model.
\newblock In {\em Proceedings of the 54th Annual Meeting of the Association for
  Computational Linguistics (ACL)}, pages 994--1003, Berlin, Germany, Aug.
  2016. Association for Computational Linguistics.

\bibitem{li2017learning}
Tianye Li, Timo Bolkart, Michael~J Black, Hao Li, and Javier Romero.
\newblock Learning a model of facial shape and expression from 4d scans.
\newblock {\em ACM Trans. Graph.}, 36(6):194--1, 2017.

\bibitem{ex10}
Yunlong {Liang}, Fandong {Meng}, Jinan {Xu}, Yufeng {Chen}, and Jie {Zhou}.
\newblock {MSCTD: A Multimodal Sentiment Chat Translation Dataset}.
\newblock {\em arXiv e-prints}, page arXiv:2202.13645, Feb. 2022.

\bibitem{ex18}
Pierre-Emmanuel Mazar{\'e}, Samuel Humeau, Martin Raison, and Antoine Bordes.
\newblock Training millions of personalized dialogue agents.
\newblock In {\em Proceedings of the 2018 Conference on Empirical Methods in
  Natural Language Processing (EMNLP)}, pages 2775--2779, Brussels, Belgium,
  Oct.-Nov. 2018. Association for Computational Linguistics.

\bibitem{ex11}
Arsha Nagrani, Joon~Son Chung, Weidi Xie, and Andrew Zisserman.
\newblock Voxceleb: Large-scale speaker verification in the wild.
\newblock {\em Computer Speech \& Language}, 60:101027, 2020.

\bibitem{ex36}
Adam Paszke, Sam Gross, Francisco Massa, Adam Lerer, James Bradbury, Gregory
  Chanan, Trevor Killeen, Zeming Lin, Natalia Gimelshein, Luca Antiga, Alban
  Desmaison, Andreas K\"{o}pf, Edward Yang, Zach DeVito, Martin Raison, Alykhan
  Tejani, Sasank Chilamkurthy, Benoit Steiner, Lu Fang, Junjie Bai, and Soumith
  Chintala.
\newblock {\em PyTorch: An Imperative Style, High-Performance Deep Learning
  Library}.
\newblock Curran Associates Inc., Red Hook, NY, USA, 2019.

\bibitem{pataranutaporn2021ai}
Pat Pataranutaporn, Valdemar Danry, Joanne Leong, Parinya Punpongsanon, Dan
  Novy, Pattie Maes, and Misha Sra.
\newblock Ai-generated characters for supporting personalized learning and
  well-being.
\newblock {\em Nature Machine Intelligence}, 3(12):1013--1022, 2021.

\bibitem{poole2022dreamfusion}
Ben Poole, Ajay Jain, Jonathan~T Barron, and Ben Mildenhall.
\newblock Dreamfusion: Text-to-3d using 2d diffusion.
\newblock {\em arXiv preprint arXiv:2209.14988}, 2022.

\bibitem{popov2021grad}
Vadim Popov, Ivan Vovk, Vladimir Gogoryan, Tasnima Sadekova, and Mikhail
  Kudinov.
\newblock Grad-tts: A diffusion probabilistic model for text-to-speech.
\newblock In {\em International Conference on Machine Learning (ICML)}, pages
  8599--8608. PMLR, 2021.

\bibitem{ex6}
Soujanya Poria, Devamanyu Hazarika, Navonil Majumder, Gautam Naik, Erik
  Cambria, and Rada Mihalcea.
\newblock {MELD}: A multimodal multi-party dataset for emotion recognition in
  conversations.
\newblock In {\em Proceedings of the 57th Annual Meeting of the Association for
  Computational Linguistics (ACL)}, pages 527--536, Florence, Italy, July 2019.
  Association for Computational Linguistics.

\bibitem{ex39}
Matt Post.
\newblock A call for clarity in reporting {BLEU} scores.
\newblock In {\em Proceedings of the Third Conference on Machine Translation:
  Research Papers}, pages 186--191, Belgium, Brussels, Oct. 2018. Association
  for Computational Linguistics.

\bibitem{ramesh2022hierarchical}
Aditya Ramesh, Prafulla Dhariwal, Alex Nichol, Casey Chu, and Mark Chen.
\newblock Hierarchical text-conditional image generation with clip latents.
\newblock {\em arXiv preprint arXiv:2204.06125}, 2022.

\bibitem{ramesh2021zero}
Aditya Ramesh, Mikhail Pavlov, Gabriel Goh, Scott Gray, Chelsea Voss, Alec
  Radford, Mark Chen, and Ilya Sutskever.
\newblock Zero-shot text-to-image generation.
\newblock In {\em International Conference on Machine Learning (ICML)}, pages
  8821--8831. PMLR, 2021.

\bibitem{ex32}
Yi Ren, Chenxu Hu, Xu Tan, Tao Qin, Sheng Zhao, Zhou Zhao, and Tie-Yan Liu.
\newblock Fastspeech 2: Fast and high-quality end-to-end text to speech, 2020.

\bibitem{ren2019fastspeech}
Yi Ren, Yangjun Ruan, Xu Tan, Tao Qin, Sheng Zhao, Zhou Zhao, and Tie-Yan Liu.
\newblock Fastspeech: Fast, robust and controllable text to speech.
\newblock {\em Advances in Neural Information Processing Systems}, 32, 2019.

\bibitem{rombach2021highresolution}
Robin Rombach, Andreas Blattmann, Dominik Lorenz, Patrick Esser, and Björn
  Ommer.
\newblock High-resolution image synthesis with latent diffusion models, 2021.

\bibitem{saharia2022photorealistic}
Chitwan Saharia, William Chan, Saurabh Saxena, Lala Li, Jay Whang, Emily
  Denton, Seyed Kamyar~Seyed Ghasemipour, Burcu~Karagol Ayan, S~Sara Mahdavi,
  Rapha~Gontijo Lopes, et~al.
\newblock Photorealistic text-to-image diffusion models with deep language
  understanding.
\newblock {\em arXiv preprint arXiv:2205.11487}, 2022.

\bibitem{siarohin2019first}
Aliaksandr Siarohin, St{\'e}phane Lathuili{\`e}re, Sergey Tulyakov, Elisa
  Ricci, and Nicu Sebe.
\newblock First order motion model for image animation.
\newblock {\em Advances in Neural Information Processing Systems}, 32, 2019.

\bibitem{singer2022make}
Uriel Singer, Adam Polyak, Thomas Hayes, Xi Yin, Jie An, Songyang Zhang, Qiyuan
  Hu, Harry Yang, Oron Ashual, Oran Gafni, et~al.
\newblock Make-a-video: Text-to-video generation without text-video data.
\newblock {\em arXiv preprint arXiv:2209.14792}, 2022.

\bibitem{JMLR:v9:vandermaaten08a}
Laurens van~der Maaten and Geoffrey Hinton.
\newblock Visualizing data using t-sne.
\newblock {\em Journal of Machine Learning Research}, 9(86):2579--2605, 2008.

\bibitem{vishnubhotla-etal-2019-fictional}
Krishnapriya Vishnubhotla, Adam Hammond, and Graeme Hirst.
\newblock Are fictional voices distinguishable? classifying character voices in
  modern drama.
\newblock In {\em Proceedings of the 3rd Joint {SIGHUM} Workshop on
  Computational Linguistics for Cultural Heritage, Social Sciences, Humanities
  and Literature}, pages 29--34, Minneapolis, USA, June 2019. Association for
  Computational Linguistics.

\bibitem{VM2022-TED}
Krishnapriya Vishnubhotla and Saif~M. Mohammad.
\newblock Tweet emotion dynamics: Emotion word usage in tweets from us and
  canada.
\newblock In {\em Proceedings of the Thirteenth International Conference on
  Language Resources and Evaluation (LREC 2022)}, Marseille, France, 2022.

\bibitem{vougioukas2020realistic}
Konstantinos Vougioukas, Stavros Petridis, and Maja Pantic.
\newblock Realistic speech-driven facial animation with gans.
\newblock {\em International Journal of Computer Vision}, 128(5):1398--1413,
  2020.

\bibitem{ex12}
Kaisiyuan Wang, Qianyi Wu, Linsen Song, Zhuoqian Yang, Wayne Wu, Chen Qian, Ran
  He, Yu Qiao, and Chen~Change Loy.
\newblock Mead: A large-scale audio-visual dataset for emotional talking-face
  generation.
\newblock In Andrea Vedaldi, Horst Bischof, Thomas Brox, and Jan-Michael Frahm,
  editors, {\em Computer Vision -- ECCV 2020}, pages 700--717, Cham, 2020.
  Springer International Publishing.

\bibitem{Wang2020encoding}
Zhe Wang, Chun-Hua Wu, Qing-Biao Li, Bo Yan, and Kang-Feng Zheng.
\newblock Encoding text information with graph convolutional networks for
  personality recognition.
\newblock {\em Applied Sciences}, 10(12), 2020.

\bibitem{wu2021imitating}
Haozhe Wu, Jia Jia, Haoyu Wang, Yishun Dou, Chao Duan, and Qingshan Deng.
\newblock Imitating arbitrary talking style for realistic audio-driven talking
  face synthesis.
\newblock In {\em Proceedings of the 29th ACM International Conference on
  Multimedia (ACMMM)}, pages 1478--1486, 2021.

\bibitem{ex21}
Yuwei Wu, Xuezhe Ma, and Diyi Yang.
\newblock Personalized response generation via generative split memory network.
\newblock In {\em Proceedings of the 2021 Conference of the North American
  Chapter of the Association for Computational Linguistics: Human Language
  Technologies(ACL-HLT)}, pages 1956--1970, Online, June 2021. Association for
  Computational Linguistics.

\bibitem{ex17}
Yujie Xing and Raquel Fern{\'a}ndez.
\newblock Automatic evaluation of neural personality-based chatbots.
\newblock In {\em Proceedings of the 11th International Conference on Natural
  Language Generation (NLG)}, pages 189--194, Tilburg University, The
  Netherlands, Nov. 2018. Association for Computational Linguistics.

\bibitem{ex15}
Junichi Yamagishi, Christophe Veaux, and Kirsten MacDonald.
\newblock Cstr vctk corpus: English multi-speaker corpus for cstr voice cloning
  toolkit (version 0.92).
\newblock \url{https://doi.org/10.7488/ds/2645.}, 2019.

\bibitem{ex3}
Wenmeng Yu, Hua Xu, Fanyang Meng, Yilin Zhu, Yixiao Ma, Jiele Wu, Jiyun Zou,
  and Kaicheng Yang.
\newblock {CH}-{SIMS}: A {C}hinese multimodal sentiment analysis dataset with
  fine-grained annotation of modality.
\newblock In {\em Proceedings of the 58th Annual Meeting of the Association for
  Computational Linguistics (ACL)}, pages 3718--3727, Online, July 2020.
  Association for Computational Linguistics.

\bibitem{zadeh2018multi}
Amir Zadeh, Paul~Pu Liang, Soujanya Poria, Prateek Vij, Erik Cambria, and
  Louis-Philippe Morency.
\newblock Multi-attention recurrent network for human communication
  comprehension.
\newblock In {\em Thirty-Second AAAI Conference on Artificial Intelligence
  (AAAI)}, 2018.

\bibitem{ex1}
Amir Zadeh, Rowan Zellers, Eli Pincus, and Louis-Philippe Morency.
\newblock Multimodal sentiment intensity analysis in videos: Facial gestures
  and verbal messages.
\newblock {\em IEEE Intelligent Systems}, 31(6):82--88, 2016.

\bibitem{ex41}
Yizhe Zhang, Michel Galley, Jianfeng Gao, Zhe Gan, Xiujun Li, Chris Brockett,
  and Bill Dolan.
\newblock Generating informative and diverse conversational responses via
  adversarial information maximization.
\newblock In {\em NeurIPS 2018}, December 2018.

\bibitem{ex31}
Yizhe Zhang, Siqi Sun, Michel Galley, Yen-Chun Chen, Chris Brockett, Xiang Gao,
  Jianfeng Gao, Jingjing Liu, and Bill Dolan.
\newblock {DIALOGPT} : Large-scale generative pre-training for conversational
  response generation.
\newblock In {\em Proceedings of the 58th Annual Meeting of the Association for
  Computational Linguistics (ACL): System Demonstrations}, pages 270--278,
  Online, July 2020. Association for Computational Linguistics.

\bibitem{ex20}
Yinhe Zheng, Guanyi Chen, Minlie Huang, Song Liu, and Xuan Zhu.
\newblock Personalized dialogue generation with diversified traits.
\newblock {\em CoRR}, abs/1901.09672, 2019.

\bibitem{ex19}
Peixiang Zhong, Chen Zhang, Hao Wang, Yong Liu, and Chunyan Miao.
\newblock Towards persona-based empathetic conversational models.
\newblock In {\em Proceedings of the 2020 Conference on Empirical Methods in
  Natural Language Processing (EMNLP)}, pages 6556--6566, Online, Nov. 2020.
  Association for Computational Linguistics.

\bibitem{ex35}
Hang Zhou, Yasheng Sun, Wayne Wu, Chen~Change Loy, Xiaogang Wang, and Ziwei
  Liu.
\newblock Pose-controllable talking face generation by implicitly modularized
  audio-visual representation.
\newblock In {\em 2021 IEEE/CVF Conference on Computer Vision and Pattern
  Recognition (CVPR)}, pages 4174--4184, 2021.

\end{thebibliography}
}

\end{document}